\documentclass[conference]{IEEEtran}
\IEEEoverridecommandlockouts
\usepackage{cite}
\usepackage{amsmath,amssymb,amsfonts}
\usepackage{algorithmic}
\usepackage{graphicx}
\usepackage{textcomp}
\usepackage{xcolor}
\def\BibTeX{{\rm B\kern-.05em{\sc i\kern-.025em b}\kern-.08em
    T\kern-.1667em\lower.7ex\hbox{E}\kern-.125emX}}

\usepackage{comment}
\usepackage{soul}

\usepackage{enumitem}
\usepackage{svg}
\usepackage{subcaption}
\setstcolor{blue}
 
\newif\ifshow 
\showfalse 

\ifshow
  \includecomment{wrap}
\else
  \excludecomment{wrap} 
\fi

\begin{document}

\title{CIA: Controllable Image Augmentation Framework Based on Stable Diffusion}

\author{

\IEEEauthorblockN{Mohamed Benkedadra}
\IEEEauthorblockA{ \textit{ University of Mons } \\
Mons, Belgium \\
mohamed.benkedadra@umons.ac.be } \\

\IEEEauthorblockN{Natarajan Chidambaram}
\IEEEauthorblockA{ \textit{ University of Mons } \\
Mons, Belgium \\
natarajan.chidambaram@umons.ac.be }\\

\IEEEauthorblockN{Matei Mancas}
\IEEEauthorblockA{ \textit{ University of Mons } \\
Mons, Belgium \\
Matei.MANCAS@umons.ac.be }

\and
\IEEEauthorblockN{Dany Rimez}
\IEEEauthorblockA{ \textit{ UCLouvain } \\
Louvain-La-Neuve, Belgium \\
dany.rimez@uclouvain.be }\\

\IEEEauthorblockN{Hamed Razavi Khosroshahi}
\IEEEauthorblockA{ \textit{ Université libre de Bruxelles } \\
Brussels, Belgium \\
hamed.razavi.khosroshahi@ulb.be } \\

\IEEEauthorblockN{Benoit Macq }
\IEEEauthorblockA{ \textit{ UCLouvain } \\
Louvain-La-Neuve, Belgium \\
benoit.macq@uclouvain.be }

\and

\IEEEauthorblockN{Tiffanie Godelaine}
\IEEEauthorblockA{ \textit{ UCLouvain  } \\
Louvain-La-Neuve, Belgium \\
tiffanie.godelaine@uclouvain.be } \\

\IEEEauthorblockN{Horacio Tellez}
\IEEEauthorblockA{ \textit{ Multitel } \\
Mons, Belgium \\
hatellezp@gmail.com }\\

\IEEEauthorblockN{Sidi Ahmed Mahmoudi }
\IEEEauthorblockA{ \textit{ University of Mons } \\
Mons, Belgium \\
sidi.mahmoudi@umons.ac.be }

}

\maketitle

\begin{abstract}
Computer vision tasks such as object detection and segmentation rely on the availability of extensive, accurately annotated datasets. In this work, We present CIA, a modular pipeline, for (1) generating synthetic images for dataset augmentation using Stable Diffusion, (2) filtering out low quality samples using defined quality metrics, (3) forcing the existence of specific patterns in generated images using accurate prompting and ControlNet. In order to show how CIA can be used to search for an optimal augmentation pipeline of training data, we study human object detection in a data constrained scenario, using YOLOv8n on COCO and Flickr30k datasets. We have recorded significant improvement using CIA-generated images, approaching the performances obtained when doubling the amount of real images in the dataset. Our findings suggest that our modular framework can significantly enhance object detection systems, and make it possible for future research to be done on data-constrained scenarios. The framework is available at: github.com/multitel-ai/CIA. 
\end{abstract}

    
\section{Introduction}
\label{sec:intro}

\begin{figure}[!ht]
    \centering
    \includegraphics[width=1\linewidth]{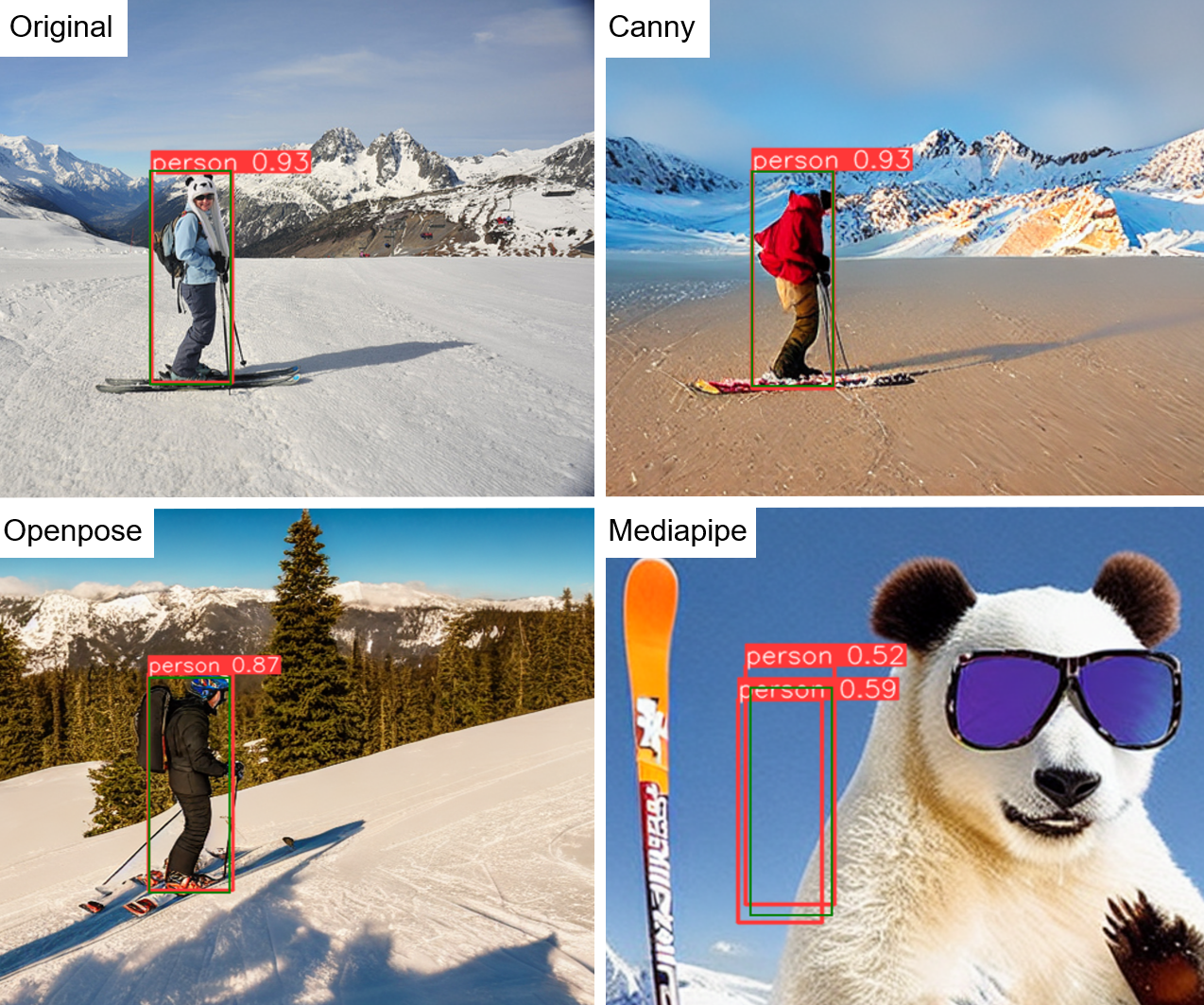}
    \caption{CIA-generated images from an image taken from the COCO dataset for different ControlNets. Either efficient (\textit{Openpose}, \textit{Canny Edge}) or inefficient (\textit{Mediapipe}) for an object-detection task. Prediction of YOLOv8n trained on the dataset corresponding to the image is shown in red, and ground truth in green.}
    \label{fig:SDCN-examples}
\end{figure}

The performance of deep learning models is dependent on the quality and diversity of the dataset they were trained on. 

Unfortunately, the creation of such high-quality and accurately annotated datasets is often challenged by the scarcity of data and the substantial costs associated with annotation \cite{su2012crowdsourcing}, especially in specialized and evolving Computer Vision tasks. Hence, other strategies are commonly used to enhance dataset quality, like Active Learning \cite{alpaper} and Data Augmentation methods \cite{XU2023109347} (image rotation, flipping, color adjustment, etc).

However, these methods modify images with simple and often content-agnostic transformations, limiting their ability to introduce completely new information into the dataset. This limitation led us to the exploration of Generative AI models like Stable Diffusion \cite{rombach2022high} that can generate entirely new images. Through the usage of ControlNet \cite{zhang2023adding} with predefined features extracted from the original image, we can tailor the generation process to meet specific task requirements. This creates an unprecedented opportunity to augment datasets beyond traditional methods. Concurrently, when dealing with synthetic data augmentation, we want to generate the most useful images for model training. This raises the question of how to assess the quality and relevance of the generated data. Finally, a pivotal question arises: 

Can the quality of region of interest vision datasets be enhanced to ensure better model performance through the incorporation of images generated with Controlled Stable Diffusion ?

To answer that question, this work introduces CIA, a modular framework for data augmentation. It integrates Stable Diffusion with Control Net models and is able to :

\begin{enumerate}[label=(\arabic*)]
    \item Generate synthetic images for dataset augmentation using both generative and classic data augmentation methods.
    \item Filter out low quality samples with defined metrics.
    \item Control the generative process to create specific patterns in the generated images for region dependent computer vision tasks (object detection, segmentation, etc.)
     \item Easily preform parallel training, testing, and comparison of multiple augmentation methods
\end{enumerate}

We prove the efficacy of CIA, through a case study on human object detection, in a scenario where the low amount of data severely limits the performances of the trained model. Examples of images generated with the proposed framework can be seen in Fig.\ref{fig:SDCN-examples}. 
\section{Related Works}
\label{sec:rel_works}

Data augmentation has become an indispensable strategy for enhancing the quality and diversity of visual datasets and improving models' performances in Computer Vision tasks. Shorten and Khoshgoftaar's comprehensive review \cite{shorten2019survey} extensively explores the diverse range of techniques employed, spanning from fundamental geometric transformations \cite{XU2023109347} to sophisticated generative methods such as Stable Diffusion \cite{rombach2022high} and ControlNet \cite{zhang2023adding}. 

These advanced techniques can generate novel content and scene conditions. For example, introducing new variations in weather, people position, object appearance, image style, etc. This essential for mitigating the limitations posed by inadequate datasets, ultimately enhancing the performance and reliability of models.

Chen et al. \cite{chen2021detection} preform scale-aware data augmentation strategies for region dependent tasks. They focus on adding objects of different scales through computationally efficient, zoom-in/out operations. Ghiasi et al. \cite{ghiasi2021augmentation} expanded on this approach by copy-pasting the zoomed objects at various scales into different backgrounds.

\subsection{Stable Diffusion for Data Augmentation} 

Eliassen and Ma \cite{eliassen2022data} demonstrated how Stable Diffusion, combined with Active Learning, can effectively re-balance classification datasets, notably outperforming traditional oversampling methods on CIFAR-10. Trabucco et al. \cite{trabucco2023effective} demonstrated the efficacy of text-to-image diffusion models in creating synthetic images for data augmentation. Similarly, Azizi et al. \cite{azizi2023synthetic} highlighted how synthetic data from diffusion models can enhance ImageNet \cite{deng2009imagenet} classification. By fine-tuning text-to-image models, they achieved class-conditional models with impressive fidelity. 

For region dependent tasks such as object detection and instance segmentation, Ge et al. \cite{ge2022synthesis} introduced a text-to-image synthesis paradigm leveraging DALL-E \cite{ramesh2021zero}. Their method generates diverse labeled data by utilizing segmentation masks to separately produce foregrounds (objects) and backgrounds (scenes). However, the approach exhibits limitations in the quality of the generated samples due to the artificial merging of generated objects onto backgrounds, resulting in a noticeable discontinuity between the foreground and background elements.

Wu et al. \cite{wu2023image} focused on image augmentation with Controlled Diffusion. Their method significantly boosts performance with minimal training data.

Although these studies collectively showcase the transformative impact of Stable Diffusion models, 
none of them offer a complete and reliable pipeline for generative data augmentation. They do not provide an easy to set up tool for quickly and efficiently testing the different augmentation strategies, or for employing quality metrics to evaluate synthetic data.


\subsection{Synthetic Images Quality Assessment} 

Assessing the quality of the synthesized samples presents a notable challenge. The complexity stems from the subjective and multidimensional nature of "quality", as its definition can defer depending on the intended application of the generated data. The literature often highlights the use of Image Quality Assessment (IQA) metrics for evaluating visual quality. Active Learning metrics can be used for assessing the potential impact of the data on model training.  

\subsubsection{IQA Metrics}
\label{qualitymetrics}

IQA metrics focus on different features and patterns in the image to quantify its quality. Blind/Referenceless Image Spatial Quality Evaluator (BRISQUE) \cite{mittal2011blind}, measures contrast, luminosity, distortion, etc, to quantify anomalies in generated images. The Neural Image Assessment (NIMA) model \cite{talebi2018nima} employs a CNN trained to measure the aesthetics and realism of synthetic images, and outputs a distribution of scores that represent different criteria. As introduced in \cite{wang2023exploring}, ClipIQA leverages the power of large-scale pre-trained vision-language models to predict image quality without reference images. It was trained on specific features related not only to quality but to the general look, feel, content, and context of the image.

\subsubsection{Active Learning metrics}
\label{sec:Related_work:CORE-SET}

Unlike Model-Agnostic metrics, Model-Aware quality metrics rely on the discriminative task's model for quality assessment. Active learning sampling strategies are often employed to assess the quality of the synthetic data by predicting its impact on model performance. Uncertainty based and diversity based methods \cite{alpaper} are the most common. In the example of object detection using YOLO, we can use the detection confidence score as a measure of the distance between a new synthetic image and the average distribution of real images used to train the baseline model.

\section{Proposed CIA Framework} 

\begin{figure*}[!h]
  \centering
  \includegraphics[width=1\linewidth]{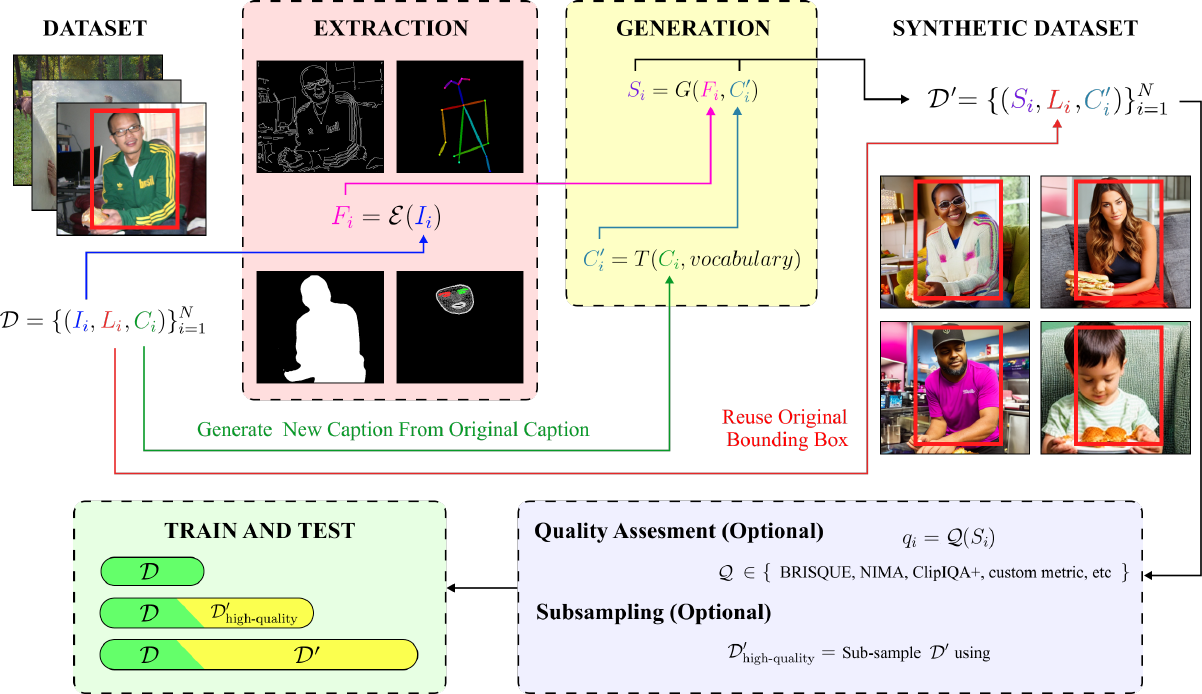}
  \caption{The CIA Framework for improving object detection accuracy through data augmentation using Stable Diffusion and ControlNet. Real images are taken from the COCO dataset. Notations used in the figure are further explained in the text.}
  \label{fig:BBOX-SDCN}
\end{figure*}

CIA is composed of four modules, as seen in Fig.\ref{fig:BBOX-SDCN}.

Initially, an \textit{Extraction} module performs feature extraction from original images, to acquire the control features that maintain the integrity of the dataset's intrinsic characteristics. These features are used in the next phase by the ControlNet to condition the output of Stable Diffusion, thus adding an extra control sequence beyond the conventional text prompt. 

The \textit{Generation} module takes in the extracted features combined with text prompts to synthesize new images. The prompts are either manually specified, or automatically generated. Optionally, In order to put constraints on the resulting dataset quality, the \textit{Quality Assessment} module can filter the generated images using chosen quality metrics, which allows for retaining only the highest-quality images. 

The final stage of the pipeline is the \textit{Train and Test} block. Through training different models, we can explore the effects of using various combinations of original and synthetic data on task performance.

\subsection{Extraction}


We begin by extracting features specific to the chosen ControlNet. Although custom extractors can be added, a few are implemented by default in CIA and cover some popular domains, from extracting human features through poses \cite{cao2017realtime} or faces \cite{grishchenko2020attention}, to broader generic features like edges \cite{canny1986computational} and segmentation masks \cite{mseddi2021fire}.


Let $ \mathcal{D} $ denote the original dataset of $ N $ real images, where each image $ I_{i} $, has a label $ L_{i} $, and a caption $ C_{i} $. Then, with the selected extractor $\mathcal{E}$, the feature image $ F_{i} $ extracted from $ I_{i} $ is given by: $ F_{i}=\mathcal{E}(I_{i})$.

\subsection{Generation}

Several generators could be obtained by combining the chosen ControlNet model with any compatible Stable Diffusion model. Once the Diffusion model is chosen, the generator $G$ is able to generate the synthetic image $S_{i}$, for each extracted feature $F_{i}$ from an image $I_{i}$, and the text prompt $C_i$ (the caption of the original image). Such that $S_{i}$ is given by $S_{i} = G(F_{i}, C_{i}^{'})$.


To introduce more diversity in generated images, we use modified captions $C_{i}^{'}$. Many methods could be used to generate prompts, such as LLMs (e.g., LLama2 ~\cite{touvron2023llama}). However, the default prompt generator \textit{T} of CIA follows a simple implementation. It takes a prompt $ C_{i} $ and a \textit{vocabulary} to produce a new prompt $ C_{i}^{'}$. For example, $ T $ modifies $C_{i} =$ \textit{a man in a red shirt}, by substituting words from a \textit{vocabulary}: \{\textit{$v0$}: [\textit{man,  woman, child}] \textit{, $v1$}: [\textit{red, black, yellow}]\}. a possible modified caption could be $ C_{i}^{'} =$ \textit{a woman in a yellow shirt}. We can generate many modified captions $ C_{ij}^{'}$, meaning $j$ possibilities of synthetic images generated from a single real image where $j \in \{0, 1, 2, \ldots, (\prod_{i=1}^{n} v_i) - 1\}$.

The new text prompts are the input prompts of Stable Diffusion. Its output is conditioned by the control features from the \textit{Extraction} module. Finally, we get the new $ \mathcal{D}' $ dataset of generated images. By default, the labels of the original images are conserved in the generated ones.



\subsection{Quality Assessor and Sampler}

To assess the quality of synthetic images in $ \mathcal{D}' $, we introduce a \textit{Quality Assessment} module that filters out low quality images. Here, the quality of $ S_{i} $ can be defined according to any metric suitable for the task. The quality score $ q_{i} $ is then computed using the selected metric $\mathcal{Q}$ from the set of Quality Metrics $Q$ such that $q_{i} = \mathcal{Q}(S_{i})$. The quality metrics implemented in CIA includes IQA and Active Learning metrics.

\subsection{Train and Test}

We can train and test multiple models for the task at hand. Through modifying generation parameters, we can choose the amount of synthetic data in this training set. Optionally, if the \textit{Quality Assessment} module is used, we can control the quality thresholds of the added synthetic data. Performances of the model are evaluated on a validation set during training, and on a test set after training. Both sets are constituted of real images only.

\section{Experimental Setup}

We preformed a case study on human object detection to prove the framework's effectiveness. In this toy example, we only have access to a limited dataset that leads to suboptimal performances. The goal is to study how to optimally improve performances, by adding CIA-generated synthetic images. The Generation parameters of Stable Diffusion were not optimized and kept constant. YOLOv8n \cite{jocher2023YOLO} was used as the object detection model. For each experiment, it was trained for 300 epochs using the training parameters from \cite{yolocfg} with the SGD optimizer. Experiments were conducted on subsets of Common Objects in Context (COCO) \cite{coco} and Flickr30k Entities \cite{flickr}. 

\subsection{Datasets} 

COCO was processed to focus on a subset of images containing only one instance of the "PERSON" class, where objects take an area between 5\% and 80\% of the image. In Flickr, objects are labelled with textual segments without consistent class annotations. We processed the textual descriptions to automatically annotate the images with the "PERSON" class. The inconsistency in Flickr's annotations provided a robust stress testing ground for CIA, simulating the variability and imperfection common in real-world datasets. Three types of training sets were created for both datasets :

\subsubsection{Baseline} Contains real images. One lower $ \mathcal{D}_{250} $ (250 images) and one upper $ \mathcal{D}_{500} $ (500 images) baselines were used as basis for comparison.

\subsubsection{Synthetic} To evaluate the impact of adding synthetic images on object detection performance, a larger synthetic dataset $D_{1250}'$ (1250 images) was generated by using five distinct auto-generated captions ($ C_1',..., C_5' $) for each sample in $\mathcal{D}_{250}$. Multiple datasets were then created with different proportions (250,500,...,1250) of synthetic images sampled from $ \mathcal{D}_{1250}'$ and added to $\mathcal{D}_{250}$.

\subsubsection{Ablation} To compare the addition of new synthetic data to simply training on real data for more epochs. We duplicated the images from $\mathcal{D}_{125}$ to obtain $ \mathcal{D}_{375}^{\text{ablation}} $, $ \mathcal{D}_{500}^{\text{ablation}} $, ..., $ \mathcal{D}_{1500}^{\text{ablation}} $.

\subsection{Experiments}
\label{sec:method}

With these datasets, three experiments presented hereafter were conducted. The first one was done using both COCO and Flickr images, while only COCO was used for the two others.

\subsubsection{ControlNet effect} 

To analyze the effects of choosing a good ControlNet that fits the task, we compared several models. Four models were chosen. Some tailored for people detection (\textit{OpenPose} or \textit{MediaPipe}), and others are more generic (\textit{Canny Edge} and \textit{Segmentation}), and suitable for various types of datasets and contexts. All four are compatible with Stable Diffusion v1.5 (runwayml/stable-diffusion-v1-5) from the Hugging Face platform. Example of CIA-generated images can be observed on Fig.\ref{fig:SDCN-examples} for the first three.



\begin{figure}[!ht]
    \begin{subfigure}{0.49\linewidth}  

        \includegraphics[width=1.\linewidth,trim={0 0.2cm 0 0.2cm}, clip]{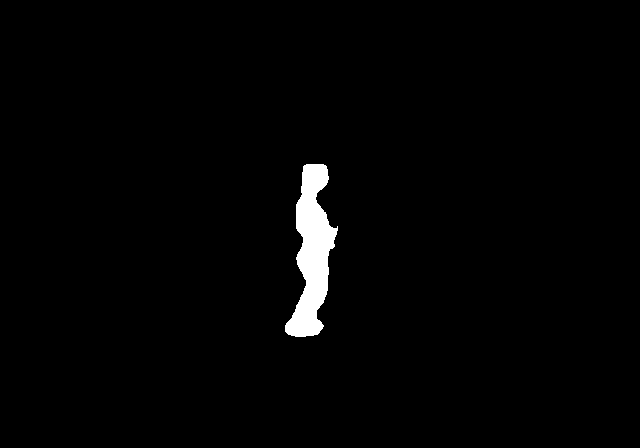} 
        \includegraphics[ width=1.\linewidth,trim={2.3cm 1.65cm 1.65cm 1.775cm}, clip]{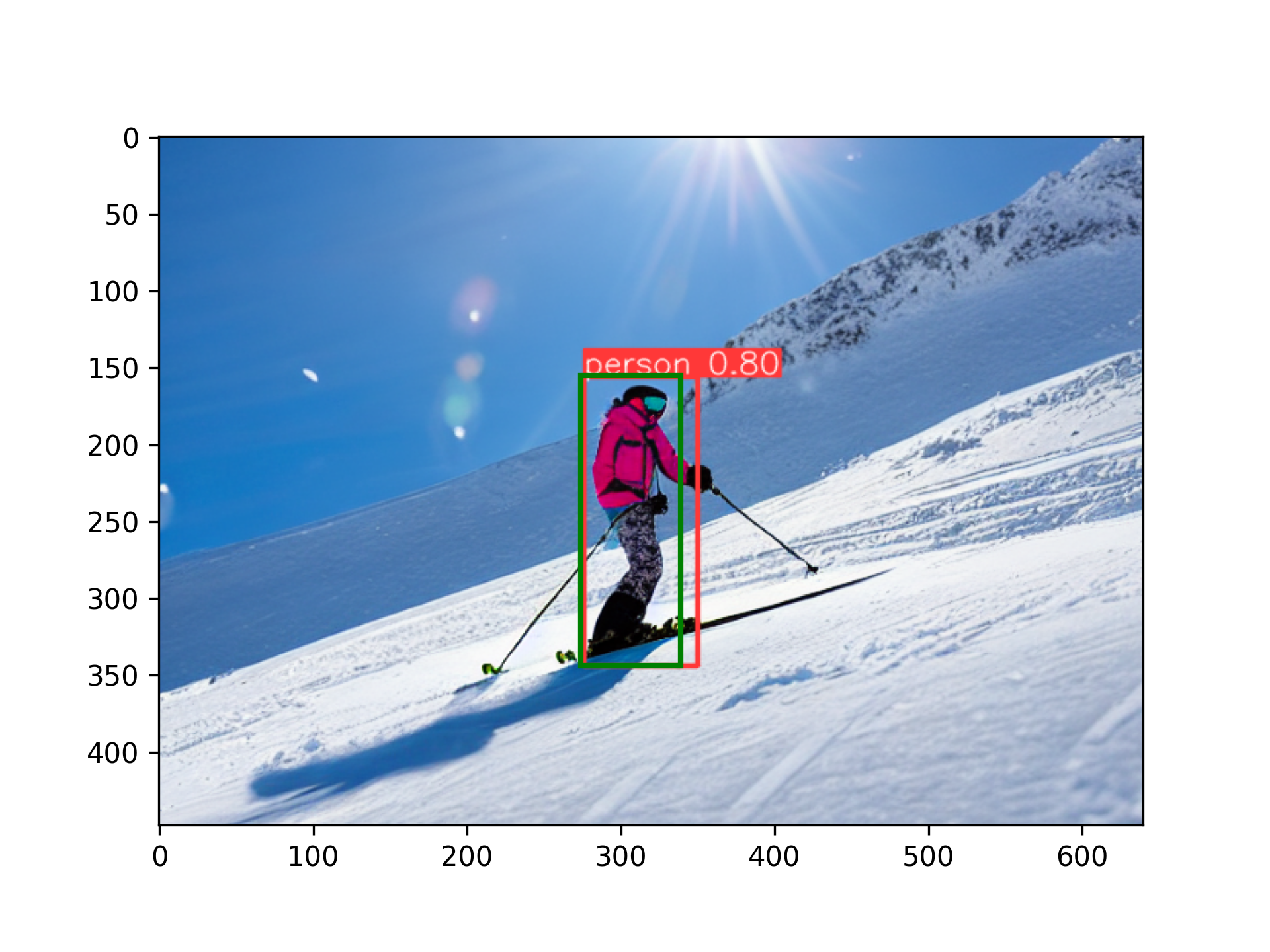} 
    \end{subfigure} 
    \begin{subfigure}{0.44\linewidth}  
        \centering
        \includegraphics[width=1.1\linewidth,trim={5.1cm 1.5cm 4.7cm 1.5cm}, clip]{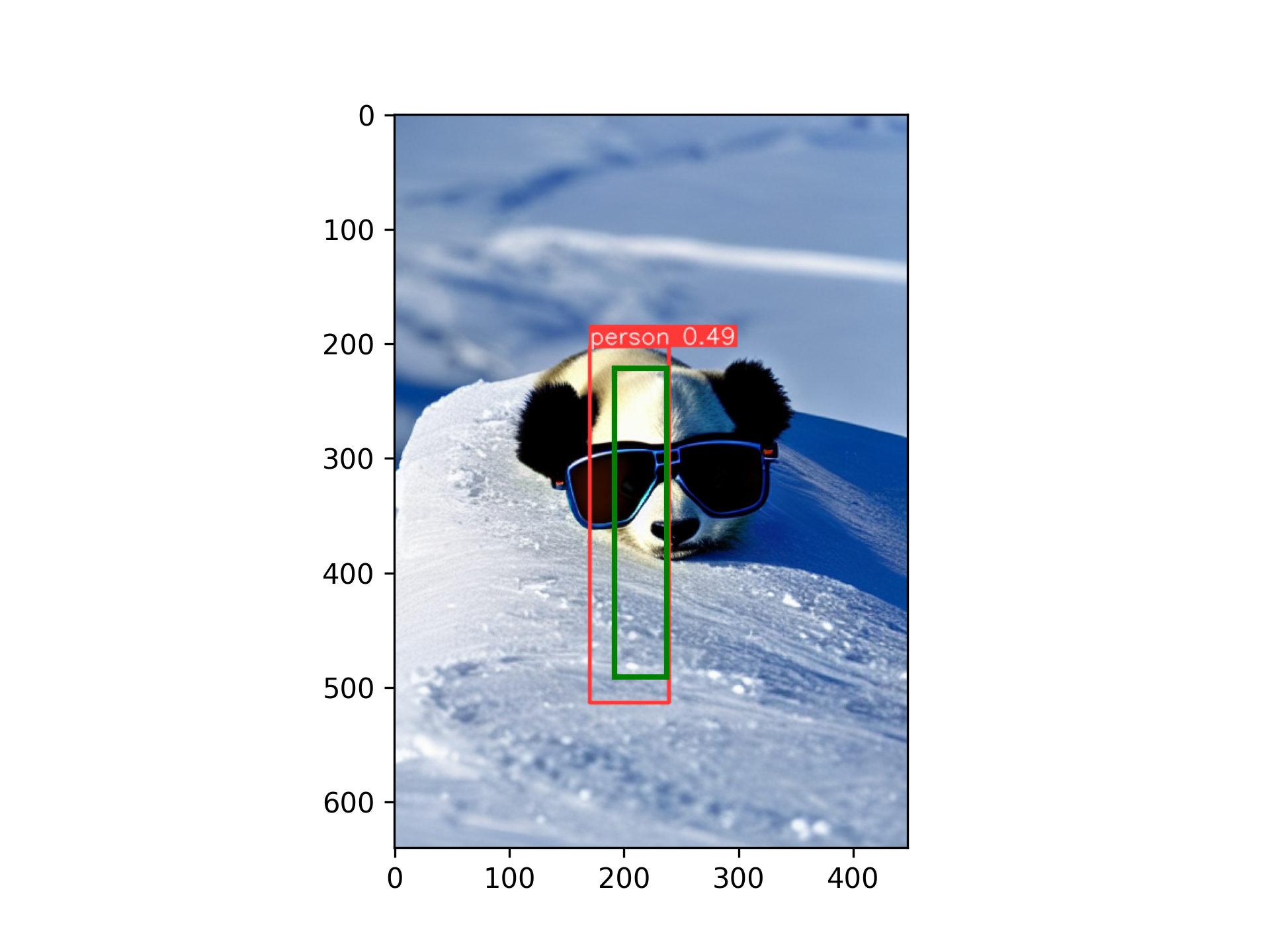} 
    \end{subfigure}
    \caption{Examples of synthetic images generated with ControlNets Segmentation and False-Segmentation from the same real image as in Fig.\ref{fig:SDCN-examples}. Left: YOLOv8m-seg's segmentation mask of the real image (top) and synthetic image generated (bottom). Right: synthetic image generated using the transposed segmentation mask.}
    \label{fig:falseseg}
\end{figure}

We added a deficient \textit{extraction} module to the case study, to understand how the usage of bad conditions affects the results. We used the \textit{Segmentation} extractor, with a transposed segmentation mask as a condition instead of the true segmentation mask. As a result (see Fig.\ref{fig:falseseg}), this new \textit{extraction} module, \textit{False-Segmentation}, generated bad quality images. Not only the shape and position of the label bounding box are affected\footnote{Bounding boxes coordinates are defined in relative coordinates and depends on the Height and Width of the image.}, but the content of the image is not necessarily coherent with the label anymore.

\begin{figure*}[!ht]
    \centering
    \includegraphics[width=0.95\textwidth]{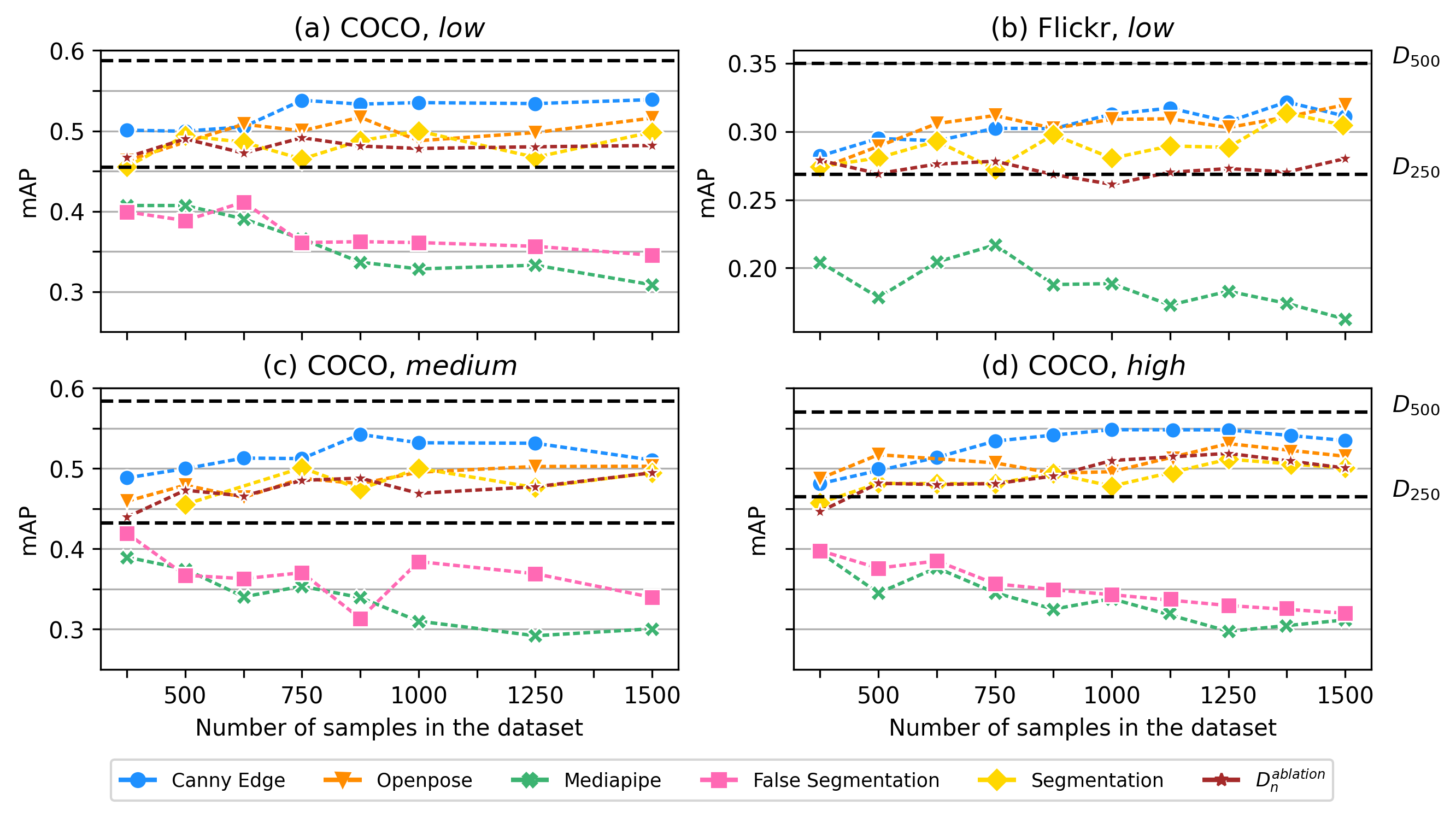}
    \caption{Performance Evaluation of the trained YOLOv8 models on test set. \textbf{Influence of 5 ControlNets} (\textit{Canny Edge}, \textit{OpenPose}, \textit{MediaPipe}, \textit{Segmentation} and \textit{False-Segmentation}) (a) on COCO dataset (b) on Flickr dataset. \textbf{Evaluation of gain using synthetic images in addition to data augmentation} on COCO dataset (c) \textit{medium} (d) \textit{high}.}
    \label{fig:results_1}
\end{figure*}

\subsubsection{Data Augmentation additivity}

This experiment aims at illustrating the first claim stated in Section \ref{sec:intro}, i.e. CIA augmentation can independently be used along with other data augmentation methods. Let's call this property additivity. We analyzed three levels of data augmentation already implemented in YOLOv8 \cite{yolodoc}. (1) Low augmentation includes scale, translation, hue saturation and mosaic. (2) Medium augmentation, adds random shear and rotation with a maximum $\pm$5 and $\pm$10 degrees respectively, and a 10\% probability of applying copy-paste. High augmentation has the same setting, with a 20\% probability of applying copy-paste and mix up. In the latter, we named the models trained on those augmentation levels: \textit{low}, \textit{medium}, and \textit{high} models. 

\subsubsection{Sampling with quality metrics}

To showcase the ability of CIA to filter images according to predefined metrics, smaller synthetic datasets $ \mathcal{D}' $ were refined. The top $ n $ images were selected from $ \mathcal{D}_{1250}^{'\mathcal{Q}} $ according to the quality metric $\mathcal{Q}$. This process enabled the creation of high-quality synthetic subsets  $ \mathcal{D}_{n-\text{high-quality}}^{'\mathcal{Q}} $. This approach yielded datasets with varying sizes from $ \mathcal{D}_{125-\text{high-quality}}^{'\mathcal{Q}} $ to $\mathcal{D}_{875-\text{high-quality}}^{'\mathcal{Q}} $, each comprising the highest quality images according to the quality metric. BRISQUE, ClipIQA, and NIMA were employed in addition to Model-Aware Active Learning metrics. Mainly, CORE-SET and confidence-score based selection. In this second method, images with the lowest confidence values predicted by the model are selected. The selection process unfolds over five rounds, with an incremental increase of 125 samples per round to produce 5 synthetic datasets $ \mathcal{D}_{n-\text{high-quality}}^{'\mathcal{Q}} $.

\section{Results} 

In this section, the results of the case study for the three aforementioned experiments are presented before being discussed to highlight the possibilities of CIA.  

\subsection{ControlNet effect}

Evaluating the influence of different ControlNets on enhancing YOLOv8's object detection capabilities, focuses on variations in mAP. This evaluation, depicted in Fig.\ref{fig:results_1} (a) and (b), reveals the significance of ControlNet choice on performance. While most ControlNets led to an increase in mAP compared to \( \mathcal{D}_{\text{250}}\), and \( \mathcal{D}^{\text{ablation}}_{n}\), none matched \( \mathcal{D}_{500}\) performance. Notably, \textit{Mediapipe} exhibits a decline in performance. This could be explained by images where the object deviates from the original bounding box (Fig.\ref{fig:SDCN-examples}). We then tested \textit{False-Segmentation}, explained in section \ref{sec:method}, and obtained similar results to \textit{Mediapipe}. This confirms that the choice of the ControlNet needs to be consistent with the task domain. 


On the contrary, \textit{Canny Edge}, \textit{OpenPose}, and \textit{Segmentation} contributed positively to mAP. This improvement was notable up to 750 synthetic samples, beyond which mAP increase was considered not significant. 





\subsection{Data augmentation additivity}

The second study aimed to analyze the impact of adding synthetic images to other data augmentation techniques to determine the value of synthetic data. The results are shown in Fig.\ref{fig:results_1} (a), (c) and (d).  We observe that using synthetic data never leads to lower performances for efficient ControlNets at all data augmentation levels, as demonstrated by the higher mAP compared to $D_{n}^{ablation}$. We can also see that the performances when using \textit{Canny Edge} with low data augmentation level are the same as the \textit{medium} baseline, coupled with the fact that \textit{high} baseline is lower than \textit{medium} baseline. Hence, classic augmentation methods are prone to cause overfitting, while CIA images guarantee good performance even at higher level of augmentation if the ControlNet is chosen correctly.


\subsection{Sampling with quality metrics}  

\begin{figure}[!h]
    \centering
    \includegraphics[width=0.95\linewidth]{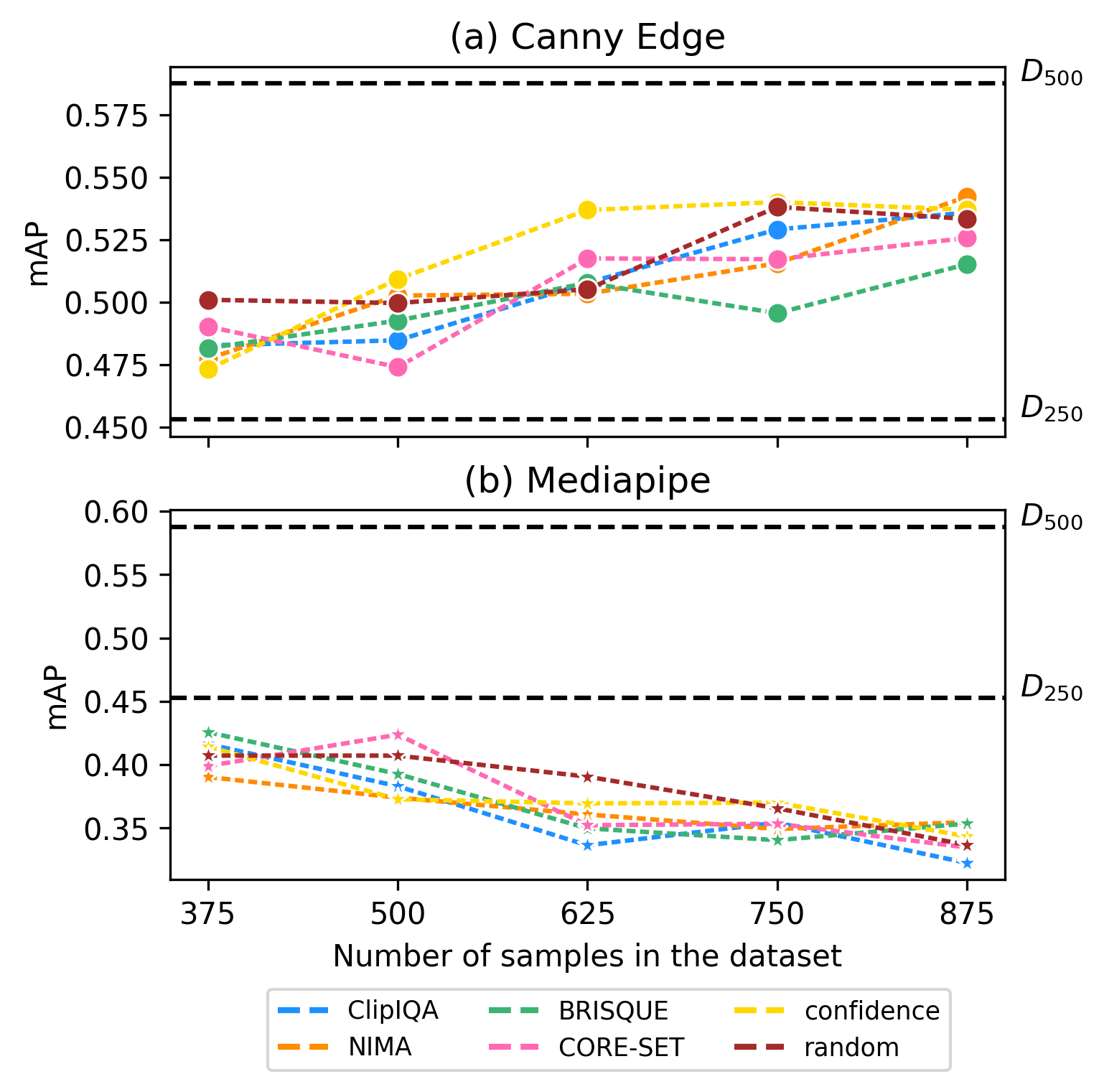}
    \caption{Performance Evaluation of the trained YOLOv8 models on test set. \textbf{Influence of sampling methods} (ClipIQA, NIMA, BRISQUE, CORE-SET, confidence) on COCO dataset for ControlNet (a) \textit{Canny Edge} (b) \textit{MediaPipe}. "random" sampling refers to plots (a) and (b) of Fig.\ref{fig:results_1} for which the synthetic images are selected randomly.}
    \label{fig:results_2}
\end{figure}

The aim of this experiment was to refine the pipeline for optimal outcomes. The results for \textit{Canny Edge} and \textit{Mediapipe} are illustrated in Fig.\ref{fig:results_2}, but results are similar for all ControlNets: none of the sampling strategies significantly outperformed random sampling. This suggests that prioritizing images based solely on features like visual quality or diversity, may not be the most effective strategy for model improvement. 

{\color{blue} 


}




\section{Discussion} 

This case study provides guidelines for using the CIA framework effectively. We demonstrated that adding synthetic images generated with the appropriate ControlNet can enhance detection performance. These images can also be used in conjunction with basic data augmentation. The analysis of the influence of sampling methods indicates that the diversity of the generated images may not be optimal. Exploring other hyperparameters during the generation process may lead to better results. Nonetheless, it is still far superior to classical methods even at very high levels of augmentation.

\begin{figure}[!t]
    \begin{subfigure}{0.495\linewidth}   
        \includegraphics[width=1.01\linewidth,trim={0 4cm 0 4cm}, clip]{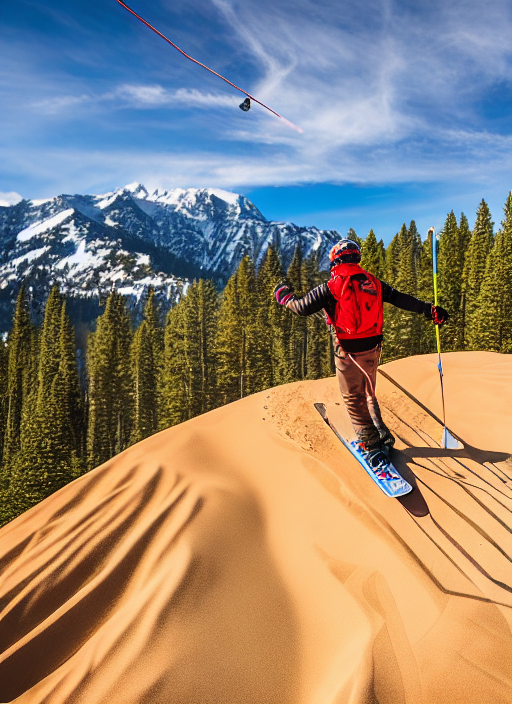}
    \end{subfigure}
    \hfill
    \begin{subfigure}{0.495\linewidth} 
        \includegraphics[width=1.01\linewidth,trim={0 4cm 0 4cm}, clip]{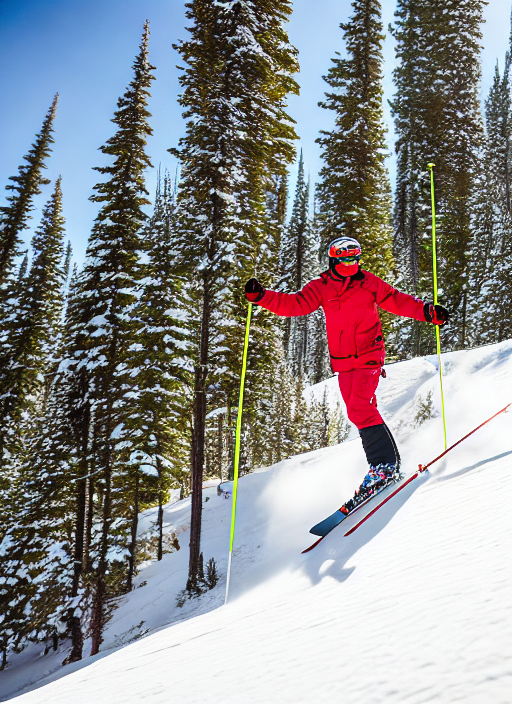} 
    \end{subfigure}
    \begin{subfigure}{0.495\linewidth}  
        \includegraphics[width=1.01\linewidth,trim={0 0.2cm 0 0.2cm}, clip]{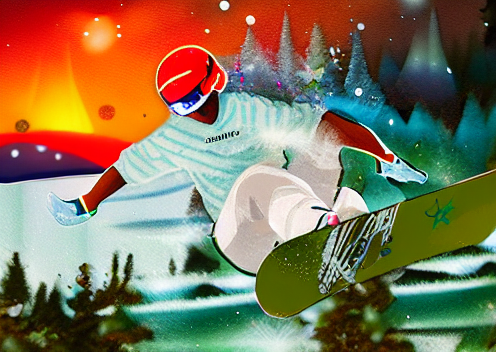} 
    \end{subfigure}
    \hfill
    \begin{subfigure}{0.495\linewidth} 
        \includegraphics[width=1.01\linewidth,trim={0 0.4cm 0 0.4cm}, clip]{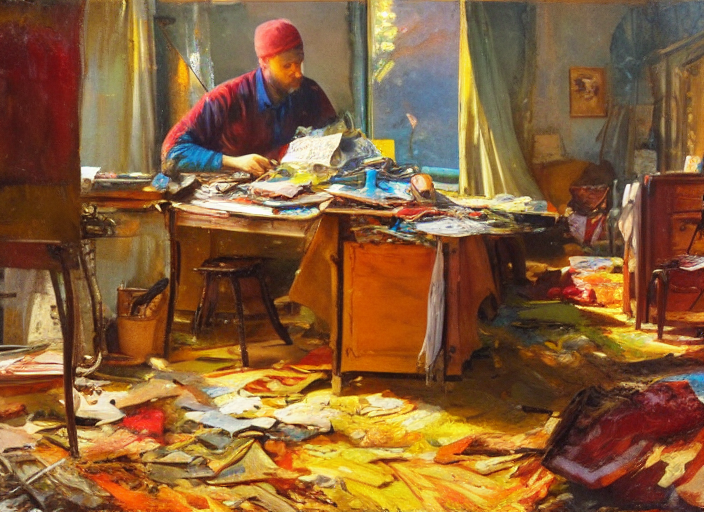} 
    \end{subfigure}
    \caption{Examples of synthetic images generated with CIA from different reference image. (Top) Changes in the point of view with the same reference image (turning back or turning away). (Bottom) Changes in style (from photography to poster or painting).}
    \label{fig:res-exmpl}
\end{figure}

An overview of the different types of images that can be produced with the five studied ControlNets was already given in Fig.\ref{fig:SDCN-examples}, Fig.\ref{fig:BBOX-SDCN} and Fig.\ref{fig:falseseg}. However, Fig.\ref{fig:res-exmpl} displays the introduction of new patterns in the images. Changes in the background (snow, forest, sand, etc.), point of view (turning back or turning away), and style (realistic, drawing, painting, photography, etc.) can be observed. Such differences could be of high interest depending on the task. This is merely a glimpse of the generation possibilities, that can be tailored through the prompt and the Stable Diffusion model choice, both of which can be easily modified with the CIA framework.

\section{Conclusion}

CIA offers a plug-and-play capability for developing, testing, and evaluating custom image generation pipelines. This framework has the potential to have a significant impact on the field of computer vision by providing researchers with a powerful tool for augmenting datasets and exploring new metrics and diffusion models. We demonstrated the capabilities of CIA in augmenting limited object detection datasets. But, the adaptability of the CIA framework allows for easy extension to other tasks like classification, segmentation or tracking. It allows for the incorporation of custom Diffusion models, ControlNet models and quality metrics to further adapt CIA to any application. Moreover, through its modularity, a module can easily be replaced or added. For example, adding other generative AI methods (not based on Stable Diffusion).

\section*{Acknowledgment}

Thanks to ARIAC/TRAIL for funding this research.


\bibliographystyle{ieeetr}
\bibliography{main}

\end{document}